\documentclass[sigconf]{acmart}

\usepackage{multirow}
\usepackage{adjustbox}
\usepackage{array}
\usepackage{mdframed}
\usepackage{enumitem}
\usepackage{amsmath}
\usepackage{subcaption}

\renewcommand\footnotetextcopyrightpermission[1]{}

\newcommand{\best}[1]{\textbf{\underline{#1}}}
\newcommand{\red}[1]{{\color{red}#1}}
\definecolor{tbl_color}{rgb}{0.94,0.94,0.94}
\definecolor{lightgray}{rgb}{0.6, 0.6, 0.6}
\newcommand{\gray}[1]{\textcolor{lightgray}{#1}}

\begin{document}

\title{Look Before You Decide: Prompting Active Deduction\\ of MLLMs for Assumptive Reasoning}

\author{Yian Li}
\affiliation{%
  \institution{Shanghai Key Lab of Intell. Info. Processing, School of CS, Fudan University, China}  
  \city{}
  \country{}}
\email{yali24@m.fudan.edu.cn}

\author{Wentao Tian}
\affiliation{%
  \institution{Shanghai Key Lab of Intell. Info. Processing, School of CS, Fudan University, China}  
  \city{}
  \country{}}
\email{wttian22@m.fudan.edu.cn}

\author{Yang Jiao}
\affiliation{%
  \institution{Shanghai Key Lab of Intell. Info. Processing, School of CS, Fudan University, China}  
  \city{}
  \country{}}
\email{yjiao23@m.fudan.edu.cn}

\author{Jingjing Chen}
\authornote{Corresponding author.}
\affiliation{%
  \institution{Shanghai Key Lab of Intell. Info. Processing, School of CS, Fudan University, China}
  \city{}
  \country{}}
\email{chenjingjing@fudan.edu.cn}

\author{Tianwen Qian}
\affiliation{%
  \institution{School of Computer Science and Technology, East China Normal University, China}  
  \city{}
  \country{}}
\email{twqian@cs.ecnu.edu.cn}

\author{Bin Zhu}
\affiliation{%
  \institution{School of Computing and Information Systems, Singapore Management University, Singapore}  
  \city{}
  \country{}}
\email{binzhu@smu.edu.sg}

\author{Na Zhao}
\affiliation{%
  \institution{Information Systems Technology and Design, Singapore University of Technology and Design, Singapore}
  \city{}
  \country{}}
\email{na_zhao@sutd.edu.sg}

\author{Yu-Gang Jiang}
\affiliation{%
  \institution{Shanghai Key Lab of Intell. Info. Processing, School of CS, Fudan University, China}  
  \city{}
  \country{}}
\email{ygj@fudan.edu.cn}

\renewcommand{\shorttitle}{Look Before You Decide: Prompting Active Deduction of MLLMs for Assumptive Reasoning}
\renewcommand{\shortauthors}{Yian Li et al.}

\begin{abstract}
    Recently, Multimodal Large Language Models (MLLMs) have achieved significant success across multiple disciplines due to their exceptional instruction-following capabilities and extensive world knowledge. However, whether these MLLMs possess human-like compositional reasoning abilities remains an open problem. To unveil their reasoning behaviors, we first curate a \textbf{M}ultimodal \textbf{A}ssumptive \textbf{R}ea\textbf{s}oning Benchmark (MARS-Bench) in this paper. Interestingly, we find that most prevalent MLLMs can be easily fooled by the introduction of a presupposition into the question, whereas such presuppositions appear naive to human reasoning. Besides, we also propose a simple yet effective method, Active Deduction (AD), a novel reinforcement learning paradigm to encourage the model to actively perform composite deduction before reaching a final decision. Equipped with the proposed AD method, a MLLM demonstrates significant improvements in assumptive reasoning abilities without compromising its general-purpose question-answering performance. We also provide extensive evaluations of both open-source and private MLLMs on MARS-Bench, along with experimental analyses of the AD method.
\end{abstract}

\ccsdesc[500]{Computing methodologies~Knowledge representation and reasoning}
\ccsdesc[300]{Computing methodologies~Computer vision}
\ccsdesc[300]{Computing methodologies~Natural language processing}

\keywords{Assumptive reasoning, MLLMs, VQA, Benchmark, GRPO}

\maketitle

\section{Introduction}
\label{sec:intro}

\begin{figure}[t]
    \centering
    \vspace{1em}
    \includegraphics[width=\linewidth]{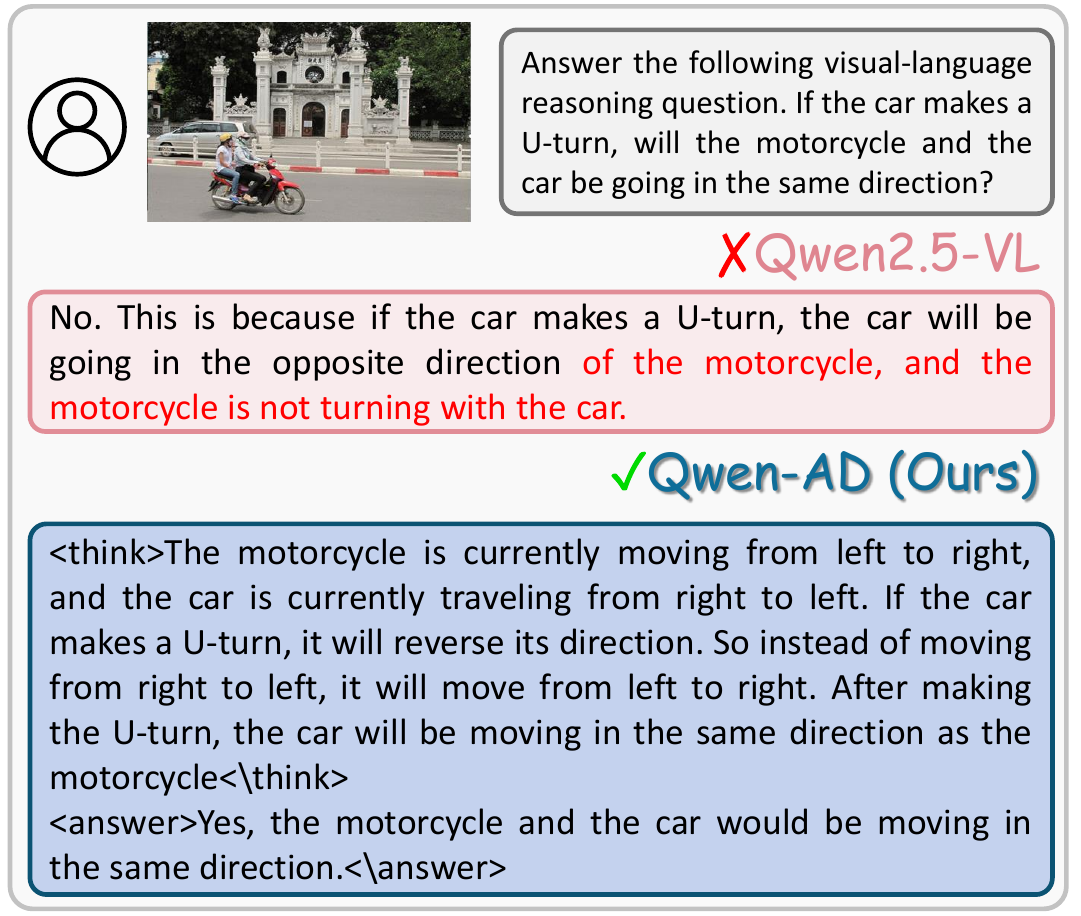}
    \caption{\textbf{The results comparison of Qwen and the Qwen equipped with our proposed Active Deduction (AD) method when faced with assumptive reasoning questions.} Qwen fails to provide the correct answer due to the plausible yet incorrect CoT it generates (highlighted in red). In contrast, our Qwen-AD can decompose the problem, deducing the answer based on the mastered knowledge. We use \red{red} to denote the wrong answer, and bulb icons to denote the CoT instructions actively generated by our model.}
    \label{fig:demo}
    \vspace{-1em}
\end{figure}

Recently, the Multimodal Large Language Models (MLLMs)~\cite{achiam2023gpt, alayrac2022flamingo, bai2023qwen, dai2024instructblip, li2023blip, li2023monkey, peng2023kosmos, jiao2024lumen, zhu2023minigpt, wang2025simplear} have been a rising research hotspot due to their potential of serving as versatile generalists across multiple disciplines. With world knowledge distilled from vast corpora, MLLMs present remarkable reasoning capabilities in solving challenging tasks. LISA~\cite{lai2024lisa} addresses the task of ``reasoning segmentation" task by integrating a MLLM with the SAM~\cite{kirillov2023segment}. Visual-CoT~\cite{shao2024visual} grounds specific image regions as intermediate steps in its reasoning process to handle complex questions. Although these MLLMs achieved unprecedented success, we wonder whether they genuinely demonstrate human-like composite reasoning steps before making the decision. 

To uncover the reasoning behaviors of MLLMs, we refactor a typical VQA sample by adding a presupposition as shown in Fig.\ref{fig:demo}. While this may seem straightforward to a human, this question can easily confuse the leading open-source MLLM, namely Qwen2.5-VL~\cite{Qwen2.5-VL}, misleading it to provide plausible yet incorrect answers. For further analysis, we provide additional guidance to the MLLM by employing the Chain-of-Thought (CoT)~\cite{wei2022chain, lu2022learn} technique, aiming to unleash its reasoning potential through multi-turn reflection. Interestingly, as demonstrated in Fig.\ref{fig:demo},  the MLLM tends to generate a specious CoT process to support its incorrect answers. 
Through the above experimental probe, it can be observed that the MLLM is prone to make decisions based on its intuition, synthesized from the knowledge stored in its memory. We call such behaviors of the MLLM as \textit{``empirical reasoning"} in this paper.

Compared to empirical reasoning, human cognition exhibits strong compositionality, allowing the expansion of new knowledge by deducing from a finite set of mastered concepts. To tackle the question in Fig.\ref{fig:demo}, it is necessary to \textbf{(1)} recognize the direction of movement of the car and motorcycle, and \textbf{(2)} comprehend the meaning of ``U-turn", finally \textbf{(3)} combine the results in (1) and (2) to reason about the ultimate car direction. However, as previously demonstrated, even the prevalent MLLM, namely Qwen2.5-VL~\cite{Qwen2.5-VL}, fails to produce these crucial reasoning steps. The underlying reason for the empirical reasoning nature of MLLMs lies in their tendency to mimic behaviors that occur with the highest probabilities across vast training data, where samples requiring complex logical reasoning are relatively scarce. 

To systematically assess the extent to which existing Multimodal Large Language Models (MLLMs) rely on empirical intuition during answer generation, we curate a novel \textbf{M}ultimodal \textbf{A}ssumptive \textbf{R}ea\textbf{s}oning Benchmark, abbreviated as MARS-Bench in this paper. In MARS-Bench, we design two sets of questions for obvious comparison. The first set of questions aims to inquire about the detailed content of the image. These questions are conventional and serve as foundational queries. In the second set of questions, we introduce a deliberately curated presupposition prior to each foundational question, imposing higher demands on the model to perform cross-referential reflection and reasoning in order to produce correct answers. By comparing the performance achieved on these two sets of questions, we can effectively examine a model's susceptibility to overreliance on its empirical intuition. Through comprehensively evaluating eight leading open-source models as well as the advanced private model, GPT-4o, on our MARS-Bench, we observe significant performance degradation across all open-source models, whereas GPT-4o demonstrates considerable robustness, which could offer promising avenues for enhancing the reasoning capabilities of existing MLLM in the future research.

To enhance logical reasoning capabilities, reasoning-oriented models in the NLP field, such as OpenAI-o1 and DeepSeek-R1, have integrated Reinforcement Learning (RL) techniques, including PPO~\cite{schulman2017proximal} and GRPO~\cite{guo2025deepseek}, which have proven highly effective. Building on this success, significant efforts have been devoted to employing Reinforcement Learning (RL) in Multimodal Large Language Models (MLLMs) to enhance reasoning capabilities in tasks such as visual counting and spatial comprehension, etc. Following this trend, we propose \textbf{Active Deduction} (AD), a novel reinforcement learning framework to enhance the MLLM's assumptive reasoning capability. Our core motivation lies in that questions of varying difficulties should be matched with corresponding levels of cognitive effort. Therefore, the proposed AD method employs a divide-and-conquer strategy in both Supervised Fine-Tuning (SFT) and RL stages. Specifically, the proposed AD method encourages the model to actively estimate the difficulty of questions. For simple questions, the model directly generates answers based on its empirical intuition, while for difficult ones, the model engages in compositional deduction before arriving at the final decision. With this dynamic adjustment feature, our AD method can significantly promote the assumptive reasoning capabilities of the existing MLLM, while preserving its general-purpose question-answering abilities. 

In general, our contributions can be summarized as follows:
\begin{itemize}
    \item We propose a novel  \textbf{M}ultimodal \textbf{A}ssumptive \textbf{R}ea\textbf{s}oning Benchmark (MARS-Bench), on which we widely assess the assumptive reasoning capabilities of prevalent open-source and private MLLMs.
    \item We introduce an Active Deduction (AD) method to enhance the existing MLLM's assumptive reasoning ability while not sacrificing its general-purpose question-answering performances.
    \item We also conduct extensive experiments and provide in-depth analyses to demonstrate the value of MARS-Bench and the effectiveness of the AD method.
\end{itemize}

\begin{figure*}[t]
  \centering
  \includegraphics[width=\linewidth]{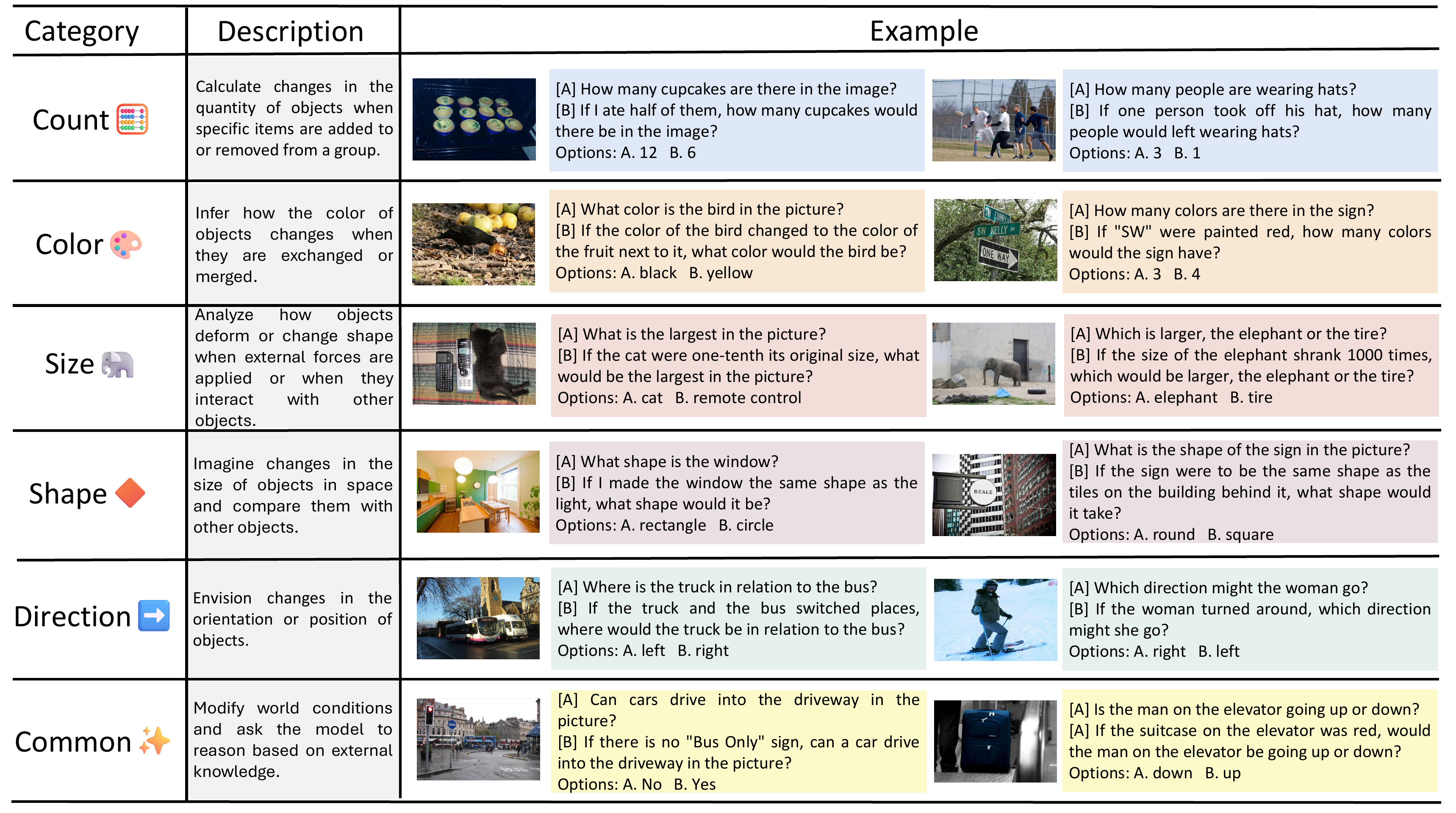}
  \caption{Demonstrations for different types of questions within our MARS-Bench. For each category, we also provide their curation rules (i.e, ``Description'') and specific examples for intuition.}
  \label{fig:exps}
\end{figure*}

\section{Related Works}
\label{sec:relate}

\textbf{Large Models with Enhanced Reasoning Capabilities.}
Recent research has increasingly focused on improving large models' reasoning abilities through various post-training approaches. Traditional methods rely on supervised fine-tuning with chain-of-thought prompting~\cite{zhang2023multimodal,lu2022learn}, which requires large amounts of high-quality annotated data. To address this limitation, reinforcement learning (RL) has emerged as a promising alternative, with methods like PPO~\cite{schulman2017proximal} and DPO~\cite{rafailov2023direct} showing success in aligning models with human preferences. A notable advancement came with Group Relative Policy Optimization (GRPO) in DeepSeekMath~\cite{shao2024deepseekmath}, which showed superior performance in mathematical reasoning and was further validated in DeepSeek-R1~\cite{guo2025deepseek}. This success has sparked a wave of GRPO applications in the multimodal domain, with recent works showing impressive results in visual-spatial reasoning~\cite{liao2025improved}, video understanding~\cite{chen2025exploring}, and visual perception tasks~\cite{liu2025visual}. Reason-RFT~\cite{tan2025reason} further demonstrates GRPO's potential in improving generalization across diverse visual reasoning tasks. However, existing approaches tend to focus on either reasoning capabilities or general-purpose functionality, making it challenging to achieve optimal performance in both aspects simultaneously. Our Active Deduction framework addresses this challenge by enabling models to dynamically adjust their reasoning process based on task complexity, maintaining both strong reasoning capabilities and general-purpose functionality.

\textbf{Benchmarks in Multimodal Comprehension Field.}
The assessment of MLLMs' reasoning capabilities has been facilitated by various benchmarks, each focusing on different aspects of multimodal understanding. Traditional benchmarks like GQA~\cite{hudson2019gqa} and OK-VQA~\cite{marino2019ok} evaluate fundamental visual reasoning and external knowledge integration. More specialized evaluations such as Science-QA~\cite{lu2022learn} and MathVista~\cite{lu2023mathvista} focus on domain-specific reasoning tasks. Comprehensive benchmarks including MME~\cite{fu2023mme} and SEED-Bench~\cite{li2024seed} assess a broader spectrum of capabilities, from commonsense reasoning to numerical calculations. However, these existing benchmarks often overlook the nuanced interplay between empirical intuition and systematic reasoning in real-world scenarios. Our proposed MARS-Bench addresses this gap by specifically evaluating models' ability to balance intuitive responses with careful analytical reasoning when faced with assumptive scenarios.

\begin{figure*}[t]
    \centering
    \begin{subfigure}[b]{0.24\textwidth}
      \includegraphics[width=\textwidth]{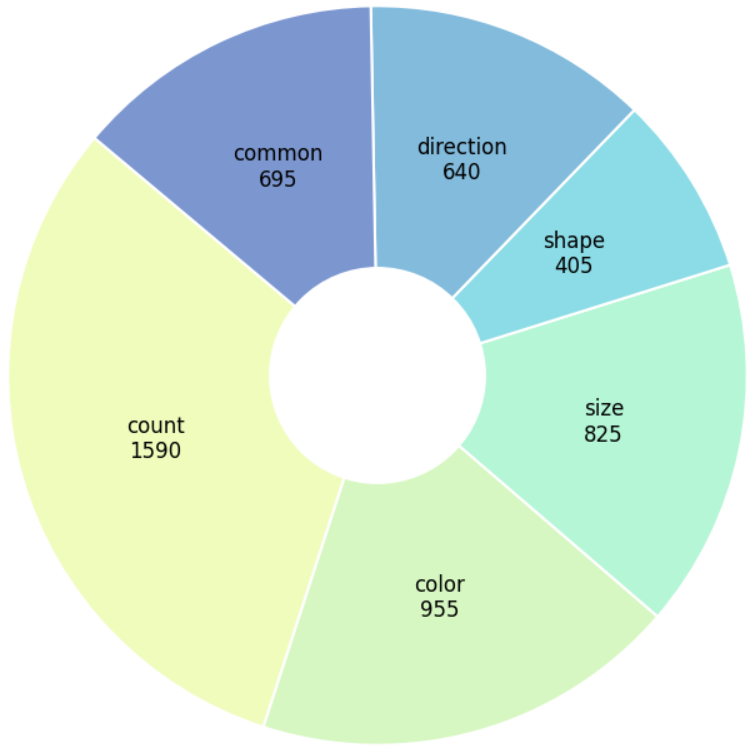}
        \caption{Question types and corresponding number.}
    \end{subfigure}
    \qquad
    \begin{subfigure}[b]{0.33\textwidth}
        \includegraphics[width=\textwidth]{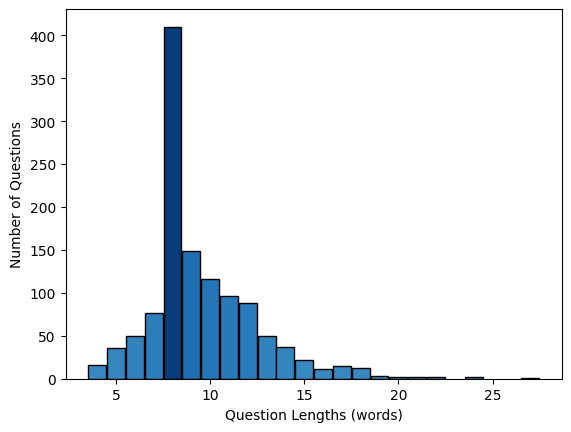}
        \caption{Distribution of word lengths in basic questions.}
    \end{subfigure}
    \qquad
    \begin{subfigure}[b]{0.33\textwidth}
        \includegraphics[width=\textwidth]{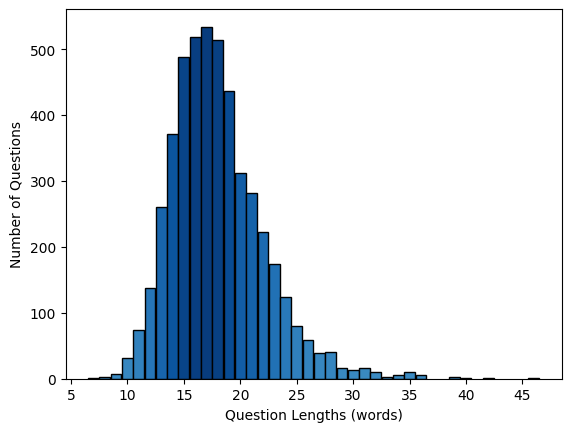}
        \caption{Distribution of word lengths in assumptive questions.}
    \end{subfigure}
    \caption{Detailed statistics of the proposed MARS-Bench. The quantitative distribution of six types of questions is shown in (a). We also list the distribution of the length of basic questions and assumptive questions in (b) and (c), respectively.}
    \label{fig:MARS}
\end{figure*}

\section{MARS-Bench}
\label{sec:mars}

This section details the MARS-Bench, a manually curated benchmark for assessing assumptive reasoning capabilities of MLLMs. We present the definition and taxonomy of assumptive questions, dataset construction, statistical analysis, and the evaluation protocol in the following.

\subsection{Problem Definition}
\textit{``Assumptive questions''} are defined as those that involve an imaginary presupposition based on known facts.
Here, ``facts'' refer to the actual information in the image, while ``presuppositions'' are hypothetical assumptions about changes to this information.
To formalize the reasoning process behind these questions, we define a function $f$: $X$→$Y$ that maps the input $x \in X$ to the output $y \in Y$ as follows:
\begin{align}
    f(v,w_{a},w_{q})=\mathop{\arg\max}\limits_{y'}\mathbf{P}(y'|v,w_{a},w_{q}).
\end{align}
where $y'$ is the output of MLLM obtained through an appropriate decoder. $v$, $w_{a}$, and $w_{q}$ represent the image, imaginary presupposition, and visual question, respectively.

We divide the assumptive questions into 6 distinct categories (count, color, shape, size, direction, and common sense) to assess the reasoning capabilities of MLLMs across multiple dimensions. Each category probes a specific aspect of reasoning, such as identifying object attributes, quantifying visual elements, or inferring spatial and directional relations. Representative examples are illustrated in Fig. \ref{fig:exps}.

\subsection{Dataset Curation}

\textbf{Data Source \& Human Annotation.}
Assumptive questions necessitate images with rich semantics to support hypothetical reasoning across various categories. To this end, we annotate images from the COCO~\cite{lin2014microsoft} validation set, which offers diverse objects and complex scenarios that closely reflect real-world distributions.

\begin{figure*}[t]
  \centering
  \includegraphics[width=\linewidth]{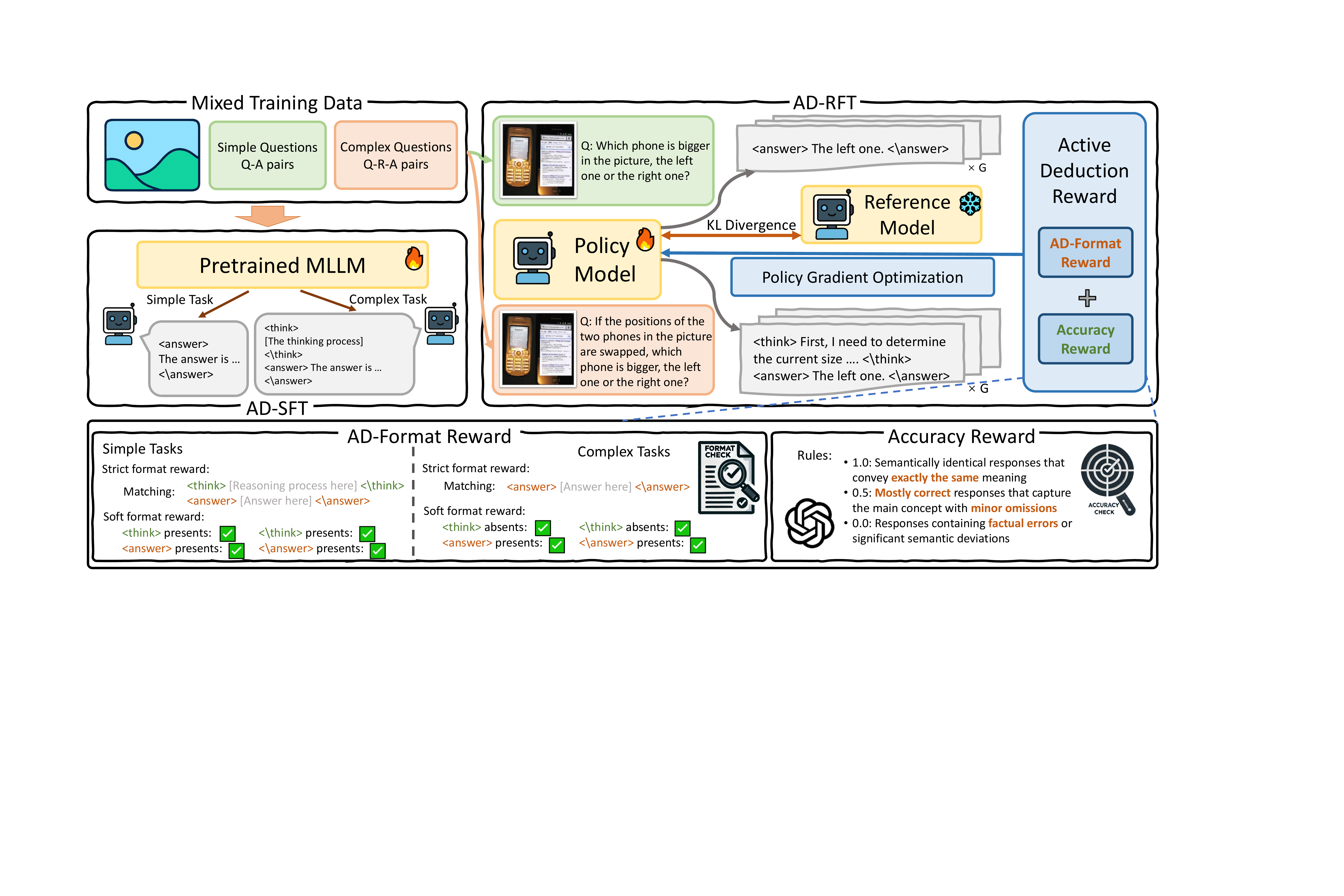}
  \caption{The overall framework of our proposed two-stage active deduction training pipeline. Our method consists of AD-SFT and AD-RFT processes. Both of them adopt a divide-and-conquer strategy to process simple and complex questions independently. We also provide illustrations for both format and accuracy rewards in our AD method for intuition.}
  \label{fig:grpo}
\end{figure*}

Our annotation involves three primary steps: (1) Annotators first determine an appropriate question type for each image, and discard images without suitable types.
(2) For each selected image, a basic visual question is crafted and then modified with a hypothetical condition to generate its assumptive counterpart.
(3) To ensure the data quality, we employ a rigorous filtering pipeline, where each question is manually verified based on two key criteria:

\begin{enumerate}[label=(\arabic*)]
    \item \textbf{Information Leakage:} Questions are removed if the answer is explicitly contained within the conditional clause (e.g., ``If I painted this bus blue, what color would it be?"), as they bypass the need for visual reasoning.
    \item \textbf{Answer Ambiguity:} We discard questions lacking sufficient visual evidence to support a reliable answer (e.g., counting objects that are partially occluded).
\end{enumerate}

\noindent\textbf{Automated Question Expansion.}
To scale our dataset, we employed GPT-4's multimodal capabilities to expand from an initial 1,200 manually annotated question-answer pairs. Given the image and corresponding annotations, GPT-4 with was prompted to generate new assumptive questions by modifying presuppositions while preserving the core question structure. As a result, the dataset was expanded to 6,000 questions. Category-specific prompts were carefully designed, and all generated samples were manually verified to ensure visual grounding, logical consistency, and answer clarity. For subsequent experiments, 880 samples were randomly sampled from the curated dataset as the training set. More details are provided in the supplementary materials.

\subsection{Dataset Statistic}

MARS-Bench comprises 1022 images, each accompanied by one basic question and four assumptive questions. The detailed distribution of different types of data is shown in Fig.~\ref{fig:MARS}(a). Besides, the lengths of questions in our MARS-Bench are shown in Fig.~\ref{fig:MARS}(b) and (c), with an average length of 9.50 and 17.92 words for basic and assumptive questions, respectively.

\subsection{Evaluation Protocol}
\label{sec:evalstr}

To facilitate quantitative evaluation, we formulate our task as binary-choice problem. However, due to the limited instruction-following capabilities of current MLLMs, models may still genetate free-from text even when explicitly prompted to choose between ``A" or ``B". Additionally, some MLLMs exhibit positional bias, tending to favor earlier options. To mitigate these issues and ensure fair evaluation, we adopt the answer ranking strategy proposed in SEED-Bench~\cite{li2024seed}. Specifically, we compute the generation loss for both candidate answers and select the one with the minimum loss as the model's prediction. Accuracy is used as our evaluation metric, where $\text{acc}_\text{b}$ and $\text{acc}_\text{a}$ represent accuracy on basic and assumptive questions, respectively.

\section{Method}
\label{sec:method}

As shown in Fig. \ref{fig:grpo}, QwenAD-Series is a novel framework designed to equip MLLMs with Active Deduction capabilities. We first describe our training data construction process, followed by the two-stage training paradigm comprising AD-SFT and AD-RFT. Finally, we elaborate on the reward design that effectively guides the model toward improved reasoning performance.

\begin{table}[t]
    \centering
    \vspace{1em}
    \begin{tabular}{p{.96\linewidth}}
    \toprule
    A conversation between User and Assistant.\\
    The user asks a question, and the assistant solves it. \textbf{If the assistant determines that the question requires multi-step reasoning or extra thinking steps}, the assistant generates a <think> tag, followed by the reasoning process enclosed within <think> </think> tags, and then provides the answer within <answer> </answer> tags, i.e., <think> reasoning process here </think> <answer> answer here </answer>. \textbf{If the question is simple and does not require additional reasoning}, the assistant directly provides the answer within <answer> </answer> tags, i.e., <answer> answer here </answer>.\\
    User: [prompt]. Assistant: \\
    \bottomrule
    \end{tabular}
    \caption{Template of the system prompt for executing our Active Deduction method. [prompt] will be replaced with the specific reasoning question during training.}
    \label{tab:prompt}
\end{table}

\subsection{Training Data Construction}

To effectively train QwenAD-Series with Active Deduction capability, we construct a comprehensive training corpus that covers both complex reasoning tasks requiring step-by-step deduction and direct visual questions solvable with minimal reasoning. The training data comprises two major components: assumptive reasoning samples from MARS-Bench and supplementary data from existing vision-language datasets.

\textbf{Assumptive Reasoning Data.} We select 704 assumptive questions from MARS-Bench as core samples of complex reasoning. GPT-4V is used to generate structured annotations that decompose each question into sequential reasoning steps. These structured annotations are then reformulated into natural language reasoning paths using an ensemble of state-of-the-art language models (GPT-4.5, GPT-4o, Claude-3.5-sonnet, and DeepSeek-V3). This process yields 2,816 diverse reasoning samples with validated reasoning paths enclosed in \texttt{<think>} tags and corresponding answers in \texttt{<answer>} tags.

\textbf{Supplementary Training Data.} To ensure model versatility, we incorporate: (1) 176 basic visual questions from MARS-Bench, (2) 3,000 complex reasoning samples from LLaVA-150K~\cite{liu2024visual}, (3) 3,000 multi-turn dialogue samples, and (4) 3,000 VQA samples from various datasets (VQA-v2~\cite{goyal2017making}, OK-VQA~\cite{marino2019ok}, GQA~\cite{hildebrandt2020scene}, etc.).

In total, we construct a comprehensive training corpus of approximately 13,000 samples, supporting both complex reasoning and general visual understanding tasks.

\subsection{Two-Stage Active Deduction Training}
\label{sec:two-stage}
We propose a two-stage training strategy to enhance the Active Deduction capabilities of our model, combining Supervised Fine-Tuning (AD-SFT) and Reinforcement Fine-Tuning (AD-RFT) to progressively refine its reasoning abilities.

\textbf{Active Deduction SFT (AD-SFT).}
In the first stage, we leverage the meticulously curated data annotated with problem complexity labels for supervised fine-tuning. For complex questions that require multi-step reasoning, reasoning paths are explicitly included within \texttt{<think>} tags. This structured annotation enables the model to develop a nuanced understanding of when and how to engage in additional multi-step reasoning. The optimization objective of AD-SFT is formulated through the following loss function:
\begin{equation}
\begin{aligned}
    \mathcal{L}_{SFT}(\theta) &= \mathbb{E}_{q, a \sim P_{\text{simple}}(Q, A)} \left( \frac{1}{|o|} \sum_{t=1}^{|o|} \log \pi_{\theta}(o_{t} \mid q, o_{<t}) \right) \\
    &\quad + \mathbb{E}_{q, r, a \sim P_{\text{complex}}(Q, R, A)} \left( \frac{1}{|o|} \sum_{t=1}^{|o|} \log \pi_{\theta}(o_{t} \mid q, o_{<t}) \right),
\end{aligned}
\end{equation}
where \( o \sim P_{\text{simple}}(Q, A) \) denotes sampling from the answer space \( A \) for simple tasks (i.e., \( o = a \)), and \( o \sim P_{\text{complex}}(Q, R, A) \) represents sampling from the concatenated sequence of reasoning steps \( R \) and answers \( A \) for complex tasks (i.e., \( o = [r, a] \)).

In the context of Active Deduction, SFT enables the model to distinguish between queries requiring multi-step reasoning and those suitable for direct answers, learning to appropriately trigger \texttt{<think>} for reasoning and \texttt{<answer>} for final responses.

\textbf{Active Deduction RFT (AD-RFT).}
In the second stage, we adopt a reinforcement learning framework based on Group Relative Policy Optimization (GRPO) \cite{shao2024deepseekmath, guo2025deepseek} to further enhance the Active Deduction capabilities of the model. Unlike conventional reinforcement learning methods such as PPO~\cite{schulman2017proximal}, that rely on an explicit critic model, AD-RFT compares multiple candidate responses within a sampled group, optimizing the model's performance through relative comparisons. This approach simplifies the optimization process and improves robustness by mitigating potential issues such as reward hacking.

\begin{table*}[t]
\centering
\scalebox{.95}{
\begin{tabular}{l|c|c|cccccccccccccc}

\toprule
\multicolumn{1}{c}{\multirow{2}{*}{Method}} & \multicolumn{1}{c}{\multirow{2}{*}{\#Para}} & \multicolumn{1}{c}{\multirow{2}{*}{P.E.}} & \multicolumn{2}{c}{Count} & \multicolumn{2}{c}{Color} & \multicolumn{2}{c}{Size} & \multicolumn{2}{c}{Shape} & \multicolumn{2}{c}{Direction} & \multicolumn{2}{c}{Common} & \multicolumn{2}{c}{Total} \\

\cline{4-17}

\multicolumn{1}{c}{} & \multicolumn{1}{c}{} & \multicolumn{1}{c}{} & $\text{acc}_\text{b}$ & $\text{acc}_\text{a}$ & $\text{acc}_\text{b}$ & $\text{acc}_\text{a}$ &  $\text{acc}_\text{b}$ & $\text{acc}_\text{a}$ &  $\text{acc}_\text{b}$ & $\text{acc}_\text{a}$ &  $\text{acc}_\text{b}$ & $\text{acc}_\text{a}$ &  $\text{acc}_\text{b}$ & $\text{acc}_\text{a}$ & $\text{acc}_\text{b}$ & $\text{acc}_\text{a}$ \\

\midrule
\multicolumn{17}{c}{\emph{Existing leading MLLMs}}\\
\midrule
\multirow{3}{*}{xGen-MM~\cite{xue2024xgen}} &
\multirow{3}{*}{4B}
& N/A & 86.2 & 73.3 & 81.7 & 70.4 & 63.0 & 56.4 & 67.9 & 71.6 & 64.1 & 52.2 & 69.8 & 61.7 & 432.7 & 385.6\\
&& ICL & 85.5 & 77.9 & 83.3 & 68.6 & 66.1 & 57.0 & 67.9 & 71.3 & 69.5 & 51.4 & 69.8 & 62.1 & 442.1 & 388.3\\
&& CoT    & 85.5 & 78.9 & 83.3 & 67.7 & 66.1 & 56.4 & 69.1 & 72.5 & 71.1 & 50.6 & 68.4 & 61.7 & 443.5 & 387.8\\
\midrule
\multirow{3}{*}{InternVL2~\cite{chen2024internvl}} &
\multirow{3}{*}{8B}
& N/A & 65.9 & 66.8 & 72.6 & 67.9 & 72.8 & 65.4 & 72.8 & 65.0 & 72.3 & 63.6 & 73.9 & 64.4 & 430.3 & 393.2 \\
&& ICL & 66.8 & 67.9 & 73.2 & 69.4 & 73.4 & 66.1 & 73.8 & 66.5 & 73.8 & 64.4 & 74.3 & 65.1 & 435.3 & 399.4 \\
&& CoT    & 67.4 & 68.3 & 74.0 & 70.5 & 74.5 & 67.0 & 74.9 & 67.3 & 74.9 & 64.8 & 75.4 & 65.8 & 441.1 & 303.7 \\
\midrule
\multirow{3}{*}{LLaVA-NeXT~\cite{liu2024llavanext}} &
\multirow{3}{*}{13B}
& N/A & 87.4 & 70.7 & 91.1 & 80.2 & 63.6 & 56.8 & 72.8 & 73.5 & 69.5 & 56.8 & 74.8 & 70.0 & 459.3 & 408.0\\
&& ICL & 81.4 & 66.0 & 89.0 & 78.5 & 65.5 & 58.6 & 75.3 & 76.2 & 71.9 & 56.4 & 73.4 & 70.9 & 456.5 & 406.8\\
&& CoT    & 81.4 & 66.9 & 88.0 & 78.9 & 66.1 & 58.0 & 74.1 & 78.1 & 67.2 & 57.2 & 72.7 & 71.4 & 449.4 & 410.6\\
\midrule
\multirow{3}{*}{LLaVA-OneVision~\cite{li2024llava-ov}} &
\multirow{3}{*}{7B}
& N/A & 86.2 & 74.4 & 92.7 & 78.9 & 67.9 & 57.4 & 75.3 & 74.7 & 72.7 & 56.8 & 74.1 & 72.5 & 468.8 & 414.7\\
&& ICL & 83.0 & 73.3 & 91.1 & 77.1 & 64.2 & 58.6 & 79.0 & 76.2 & 68.0 & 56.1 & 74.1 & 72.5 & 459.4 & 413.9\\
&& CoT    & 80.1 & 74.2 & 90.6 & 79.1 & 66.7 & 59.1 & 81.5 & 79.3 & 71.9 & 55.3 & 74.1 & 71.9 & 464.8 & 418.9\\
\midrule
\multirow{3}{*}{Qwen2.5-VL~\cite{Qwen2.5-VL}} &
\multirow{3}{*}{3B}
& N/A & 83.3 & 81.8 & 91.1 & 75.5 & 69.7 & 55.2 & 80.3 & 67.0 & 75.0 & 65.0 & 77.7 & 70.3 & 477.1 & 414.9\\
&& ICL & 81.8 & 80.1 & 91.6 & 81.2 & 75.2 & 60.2 & 80.3 & 83.0 & 70.3 & 65.4 & 80.6 & 74.6 & 479.7 & 444.5\\
&& CoT    & 81.1 & 83.3 & 90.6 & 81.8 & 74.6 & 58.8 & 82.7 & 81.2 & 68.8 & 61.9 & 80.6 & 73.9 & 478.3 & 441.0\\
\midrule
\multirow{3}{*}{Qwen2.5-VL~\cite{Qwen2.5-VL}} &
\multirow{3}{*}{7B}
& N/A & 88.7 & 84.6 & 91.1 & 78.9 & 61.8 & 60.0 & 74.1 & 66.7 & 68.8 & 61.7 & 75.5 & 66.7 & 460.0 & 418.6\\
&& ICL & 82.1 & 85.2 & 91.6 & 82.2 & 79.4 & 63.5 & 87.7 & 81.8 & 76.6 & 62.1 & 73.4 & 67.6 & 490.7 & 442.4\\
&& CoT    & 83.3 & 84.5 & \best{93.2} & 81.9 & 80.0 & 63.2 & 90.1 & 82.1 & 78.1 & 63.1 & 77.0 & 69.2 & 501.8 & 444.0\\
\midrule
GPT-4o~\cite{achiam2023gpt} & N/A & N/A & \gray{90.9} & \gray{91.7} & \gray{94.2} & \gray{87.3} & \gray{88.9} & \gray{87.4} & \gray{88.2} & \gray{80.3} & \gray{85.2} & \gray{66.0} & \gray{87.6} & \gray{73.4} & \gray{535.0} & \gray{486.0}\\
\midrule
\midrule
\multicolumn{17}{c}{\emph{Proposed Active Deduction Series}}\\
\midrule
\rowcolor{tbl_color} 
\textbf{QwenAD-SFT}        &  &  & 79.9 & 88.1 & 91.6 & 83.4 & 87.3 & 74.7 & 81.5 & 77.2 & 82.0 & 64.8 & 84.9 & 74.6 & 507.2 & 462.8 \\
\rowcolor{tbl_color} 
\textbf{QwenAD-SFT-RFT}  & 3B & AD & 87.4 & 88.4 & 90.0 & 81.5 & 87.9 & 74.7 & 85.2 & 78.1 & 80.5 & 64.5 & 84.9 & 77.3 & 515.9 & 464.5 \\
\rowcolor{tbl_color} 
\textbf{QwenAD-RFT}        &  &  & 86.5 & 88.1 & 91.1 & 78.1 & 84.9 & 73.2 & 84.0 & 77.8 & 80.5 & 60.6 & 84.2 & 75.4 & 511.0 & 453.1 \\
\midrule
\rowcolor{tbl_color} 
\textbf{QwenAD-SFT}         & &  & 84.6 & 88.7 & 88.5 & 83.5 & 87.9 & 78.6 & 85.2 & 79.0 & 82.8 & 67.6 & 89.2 & 75.5 & 518.2 & 473.0 \\
\rowcolor{tbl_color} 
\textbf{QwenAD-SFT-RFT}  & 7B & AD & \best{90.6} & \best{90.3} & 90.58 & \best{86.1} & \best{90.3} & \best{82.0} & 85.2 & 85.8 & 80.5 & \best{70.7} & \best{89.9} & 80.2 & \best{527.0} & \best{495.2} \\
\rowcolor{tbl_color} 
\textbf{QwenAD-RFT}        & &  & 88.4 & 87.2 & 88.5 & 84.0 & \best{90.3} & 72.4 & \best{90.1} & \best{86.4} & \best{83.6} & 61.5 & 84.9 & \best{80.8} & 525.8 & 472.3 \\
\bottomrule
\end{tabular}
}
\caption{
\textbf{Performance of prevalent MLLMs on six tasks within our proposed MARS-Bench.}
Here, $\text{acc}_\text{b}$ represents the accuracy for correctly answering basic questions, $\text{acc}_\text{a}$ denotes the accuracy for correctly answering assumptive questions. ``P.E." is short for ``prompt engineering" and ``AD" means using the system prompt shown in Table~\ref{tab:prompt}. We highlight the best results for open-sourced models with \best{bold}.
}
\label{tab:bench}
\end{table*}

Specifically, for a given query $q$, the current policy $\pi_{\theta_{\text{old}}}$ generates $G$ distinct candidate responses $\{o^{(1)}, o^{(2)}, ..., o^{(G)}\}$. Each response is evaluated with a task-specific reward function, yielding corresponding rewards $\{r^{(1)}, r^{(2)}, ..., r^{(G)}\}$. GRPO then normalizes these rewards to compute the relative advantage of each response as:
\begin{align}
\label{eq:advantage}
\hat{A}^{(i)} = \frac{r^{(i)} - \operatorname{mean}(\{r^{(1)}, r^{(2)}, \dots, r^{(G)}\})}{\operatorname{std}(\{r^{(1)}, r^{(2)}, \dots, r^{(G)}\})},
\end{align}
where $\hat{A}^{(i)}$ represents the normalized advantage of the $i$-th response relative to its peers. Policy updates are then performed by comparing the likelihood ratios between the new policy $\pi_{\theta}$ and the previous policy $\pi_{\theta_{\text{old}}}$. To ensure training stability and prevent excessive policy updates, we implement ratio clipping within the interval $[1-\varepsilon, 1+\varepsilon]$. Additionally, to prevent the policy from deviating too far from the reference model $\pi_{\text{ref}}$, a KL divergence penalty weighted by the coefficient $\beta$ is incorporated. The final optimization objective is formulated as:
\begin{equation}
\begin{aligned}
\label{eq:grpo-objective}
\mathcal{L}_{\text{GRPO}}(\theta) = & \mathbb{E}_{q \sim Q, \{o_i\}_{i=1}^G \sim \pi_{\theta_{\text{old}}}} \Bigg[
\frac{1}{G} \sum_{i=1}^{G} \min \Bigg( \frac{\pi_{\theta}(o_i \mid q)}{\pi_{\theta_{\text{old}}}(o_i \mid q)} \hat{A}^{(i)}, \\
&\operatorname{clip} \left( \frac{\pi_{\theta}(o_i \mid q)}{\pi_{\theta_{\text{old}}}(o_i \mid q)}, 1-\varepsilon, 1+\varepsilon \right) \cdot \hat{A}^{(i)}
\Bigg) \\
& - \beta \cdot \mathbb{D}_{\text{KL}} \left[ \pi_{\theta} ~\|~ \pi_{\text{ref}} \right]
\Bigg],
\end{aligned}
\end{equation}
where $\varepsilon$ controls the magnitude of policy updates and $\beta$ modulates the impact of the KL regularization term. The KL divergence between the learned policy $\pi_{\theta}$ and the reference policy $\pi_{\text{ref}}$ is computed as:
\begin{align}
    \mathbb{D}_{\text{KL}} \left[ \pi_{\theta} ~\|~ \pi_{\text{ref}} \right] = \frac{\pi_{\text{ref}}(o_i \mid q)}{\pi_\theta(o_i \mid q)} - \log\left(\frac{\pi_{\text{ref}}(o_i \mid q)}{\pi_\theta(o_i \mid q)}\right) - 1.
\end{align}

\subsection{Reward Design in AD-RFT}

In the AD-RFT stage, we design a composite reward mechanism that jointly evaluates semantic accuracy and format adaptation. The semantic component ensures the semantic correctness of responses, while the format component encourages dynamic output structuring based on task complexity. The final reward is computed as a weighted sum of these two components. More implementation details are in the supplementary material.

\textbf{AD Format Reward.} The format reward $r_{\text{fmt}}$ enforces adherence to the Active Deduction paradigm by evaluating the structural correctness of the model's output:
\begin{equation}
r_{\text{fmt}}(o) = \begin{cases}
r_{\text{hard}}(o) + \sum_{i=1}^{4} r_{\text{soft}}^{(i)}(o), & \text{if complex task} \\
\tilde{r}_{\text{hard}}(o) + \sum_{i=1}^{4} \tilde{r}_{\text{soft}}^{(i)}(o), & \text{otherwise}
\end{cases}
\end{equation}

The reward integrates strict (hard) and flexible (soft) matching strategies. The hard matching component $r_{\text{hard}}(o)$ grants 0.5 points for strict regex pattern matching of the complete structure, while the soft matching components $r_{\text{soft}}^{(i)}(o)$ or $\tilde{r}_{\text{soft}}^{(i)}(o)$ each grants 0.125 points for the presence of specific tags. This hybrid strategy allows the model to receive partial awards, promoting stable optimization while encouraging strict structural compliance.

For complex tasks requiring reasoning:
\begin{itemize}
    \item $r_{\text{hard}}(o)$ checks the pattern \texttt{r"$\wedge$<think>.*?</think>\textbackslash s*} \\
    \texttt{<answer>.*?</answer>"}.
    \item $r_{\text{soft}}^{(i)}(o)$ verify the presence of \texttt{<think>}, \texttt{</think>}, \texttt{<answer>}, and \texttt{</answer>} tags.
\end{itemize}

For simpler tasks:
\begin{itemize}
    \item $\tilde{r}_{\text{hard}}(o)$ checks the pattern \texttt{$\wedge$<answer>.*?</answer>}.
    \item $\tilde{r}_{\text{soft}}^{(i)}(o)$ verify the absence of \texttt{<think>} \texttt{</think>} tags and presence of \texttt{<answer>} \texttt{</answer>} tags.
\end{itemize}

\textbf{Semantic Accuracy Reward.} The semantic accuracy reward evaluates the alignment between generated and reference responses in terms of semantic similarity and factual correctness. We adopt a two-tier evaluation mechanism:
\begin{equation}
r_{\text{acc}}(o, o^*) = \begin{cases}
r_{\text{GPT}}(o, o^*), & \text{if valid response received} \\
r_{\text{SenTrans.}}(o, o^*), & \text{otherwise}
\end{cases}
\end{equation}
where $r_{\text{GPT}}(o, o^*)$ denotes the primary scoring function based on GPT-4o-mini, which serves as a semantic evaluator via carefully designed prompt template. It adopts a discrete scoring scheme with three levels:

\begin{itemize}
    \item 1.0: Exact semantic alignment with the reference
    \item 0.5: Mostly correct with minor omissions or imprecisions
    \item 0.0: Factually wrong or significantly semantically deviated
\end{itemize}

This fine-grained reward provides clear training signals and supports progressive model optimization.
To enhance robustness, a backup mechanism based on Sentence Transformers will be triggered when GPT-based evaluation is unavailable (e.g., network issues or content filtering), ensuring training continuity through similarity-based scoring.

\textbf{Reward Calculation.} The total reward for each generated response is the weighted sum of the semantic accuracy reward and the AD format reward, computed as:
\begin{align}
    r^{(i)} = \alpha \cdot r_{\text{acc}}^{(i)} + \beta \cdot r_{\text{fmt}}^{(i)},
\end{align}
where \(r_{\text{acc}}^{(i)}\) and \(r_{\text{fmt}}^{(i)}\) represent the semantic accuracy and format rewards for the \(i\)-th response, respectively, with \(\alpha\) and \(\beta\) being the corresponding coefficients for each reward.

\subsection{QwenAD-Series Models}
Based on the aforementioned SFT and RFT techniques tailored to our AD method in Sec.\ref{sec:two-stage}, we construct three variants of the QwenAD model with different configurations. Specifically, we use Qwen2.5-VL as the base model and apply SFT and RFT either individually or in combination, resulting in three distinct variants: QwenAD-SFT, QwenAD-RFT, and QwenAD-SFT-RFT. Comprehensive experimental results and in-depth comparisons are detailed in Sec.\ref{subsec:AD_eval}.

\begin{table*}[t]
\centering
\scalebox{.98}{
\begin{tabular}{l|c|ccccccc}
\toprule
\multicolumn{1}{c}{\multirow{1}{*}{Method}} & \multicolumn{1}{c}{\multirow{1}{*}{\#Para}} & MMStar & MathVista & OCRBench & SEEDBench & LLaVABench & MME & BLINK\\ 
\midrule
\multicolumn{9}{c}{\emph{Existing leading MLLMs}}\\
\midrule
InternVL2~\cite{chen2024internvl} & 8B    & 61.5 & 58.3 & 794 & 75.4 & 73.3 & 2215.1 & 50.9 \\
LLaVA-OneVision~\cite{li2024llava-ov} & 7B     & 61.9 & 62.6 & 622 & 76.7 & 81.0 & 1993.6 & 53.0 \\
LLaVA-NeXT~\cite{liu2024llavanext} & 13B  & 40.4 & 35.1 & 537 & 71.4 & 73.9 & 1745.6 & 41.2 \\
\midrule
Qwen2.5-VL~\cite{Qwen2.5-VL} & 3B           & 56.3 & 61.2 & 828 & 74.0 & 77.0 & 2199.9 & 49.1 \\
Qwen2.5-VL~\cite{Qwen2.5-VL} & 7B           & 64.1 & 68.1 & \best{888} & 77.0 & 91.0 & 2312.1 & 55.3 \\
\midrule
\midrule
\multicolumn{9}{c}{\emph{Proposed Active Deduction series}}\\
\midrule
\rowcolor{tbl_color} 
\textbf{QwenAD-SFT} &  & 55.7 & 56.8 & 765 & 73.0 & 77.7 & 2101.2 & 48.4 \\
\rowcolor{tbl_color} 
\textbf{QwenAD-SFT-RFT} & 3B & 55.9 & 58.7 & 765 & 72.7 & 73.3 & 2130.3 & 47.6 \\
\rowcolor{tbl_color} 
\textbf{QwenAD-RFT} &  & 57.1 & 59.4 & 820 & 74.2 & 78.5 & 2155.6 & 48.4 \\
\midrule
\rowcolor{tbl_color} 
\textbf{QwenAD-SFT} &  & 60.3 & 62.3 & 836 & 74.4 & 77.8 & 2197.2 & 54.4 \\
\rowcolor{tbl_color} 
\textbf{QwenAD-SFT-RFT} & 7B & 60 & 64.4 & 819 & 74.8 & 83.2 & 2132.3 & 52.9 \\
\rowcolor{tbl_color} 
\textbf{QwenAD-RFT} &  & \best{64.3} & \best{68.3} & 886 & \best{77.2} & \best{92.8} & \best{2352.6} & \best{57.2} \\
\bottomrule
\end{tabular}
}
\caption{\textbf{Results on prevalent VQA benchmarks.} We employ SEEDBench\_IMG for evaluation. All experiments, including baselines and AD methods, are conducted using VLMEvalKit~\cite{duan2024vlmevalkit}, ensuring fair and consistent comparison.}
\label{tab:vqa}
\end{table*}

\begin{table}[t]
\centering
\scalebox{.84}{
\begin{tabular}{l|cccccccccc}
\toprule
\multicolumn{1}{c}{\multirow{1}{*}{Method}} & MMStar & MME & BLINK & MARS-B & MARS-A \\ 
\midrule
Qwen2.5-VL-7B           & 64.1 & 2312.1 & 55.3 & 460.0 & 418.6 \\
\midrule
\ + vanilla reward      & 63.7 & 2351.7 & 55.3 & 503.9 & \best{475.2} \\
\ + AD format reward    & \best{64.3} & \best{2352.6} & \best{57.2} & \best{525.8} & 472.3 \\
\bottomrule
\end{tabular}
}
\caption{Comparison between our proposed AD-RFT reward and vanilla GRPO reward.}
\label{tab:abl}
\vspace{-.7em}
\end{table}

\section{Experiments and Results}
\label{sec:exp}

This section is organized as follows. In Sec.\ref{subsec:bencheval}, we  benchmark the performances of prevalent state-of-the-art MLLMs on our proposed MARS-Bench. Afterward, we introduce the implementation details and delve deeper into the proposed Qwen-AD series methods in Sec.\ref{subsec:imple_details} and Sec.\ref{subsec:AD_eval}, respectively. Finally, we further conduct comprehensive studies for in-depth analysis in Sec.\ref{subsec:compr_ana}.  

\subsection{Systematic Evaluation on MARS-Bench}
\label{subsec:bencheval}

To inspect the challenge of our MARS-Bench, we systematically evaluate a wide array of prevalent MLLMs as shown in Tab.\ref{tab:bench}. For each model, we employ tailored prompt-engineering strategies—namely In-Context Learning (ICL) and Chain-of-Thought (CoT)—to stimulate the reasoning capabilities of MLLMs. Based on Table~\ref{tab:bench}, we conduct multi-dimensional analyses as outlined below.

\noindent\textbf{Overall Performances.} In general, for basic questions ($\text{acc}_\text{b}$), most models exhibit strong performance, with Qwen2.5-VL-7B achieving the highest score of 501.8 under Chain-of-Thought (CoT) prompting. However, when confronted with assumptive questions ($\text{acc}_\text{a}$), all models show a noticeable decline in performance. This disparity underscores the greater complexity of assumptive reasoning tasks compared to basic visual questions in multimodal comprehension. Moreover, while the commercial model GPT-4o demonstrates robust and effective performance across both question types, open-source models lag considerably behind, indicating substantial room for improvement in current open-source approaches.

\noindent\textbf{Performance Breakdown.} Since different reasoning tasks exhibit varying levels of difficulties, we conduct class-wise breakdown for further analyses. Firstly, task categories like ``Color" and ``Count" require the basic perception capability, and therefore obtained higher scores across all models, with ``Color" consistently achieving above 80\% accuracy on basic questions. By contrast, ``Direction" and ``Size" tasks involve spatial reasoning and relative comparisons are more challenging, with performance dropping significantly when assumptions are involved. 
For example, even the best-performing Qwen2.5-VL-7B achieves only 63.1\% accuracy on assumptive questions in the ``Direction" category. 

\noindent\textbf{Effects of Prompt Engineering.} To unleash the reasoning potentials of MLLMs, we leverage different prompt engineering strategies including In-Context-Learning (ICL) and Chain-of-Thought (CoT). Contrary to conventional expectations, we observe that both ICL and CoT do not consistently improve assumptive reasoning performance, particularly for models with fewer than 10 billion parameters. For example, LLaVA-NeXT-13B exhibits marginal or even negative effects when these strategies are applied, with $\text{acc}_\text{a}$ dropping slightly from 408.0 (zero-shot) to 406.8 (one-shot). This indicates that the limited knowledge capacity of small models may hinder coherent chain generation, potentially introducing reasoning artifacts.

\subsection{Implementation Details of QwenAD-Series}
\label{subsec:imple_details}

We use Qwen2.5-VL 3B and 7B as our base model and conduct all experiments on 8 NVIDIA A100 GPUs using LoRA~\cite{hu2021lora} with rank 128 and AdamW optimizer. The maximum generation length is set to 2,048 tokens. For QwenAD-SFT, we train the model for 1 epoch with a learning rate of 2e-5, batch size of 128, and a warmup ratio of 0.05.
For QwenAD-RFT, the learning rate is 5e-6, with batch size of 64 and 8 candidates generated per query.
For QwenAD-SFT-RFT, the first SFT stage is the same as QwenAD-SFT. In the subsequent RFT stage, we use a learning rate of 2e-6, and other settings align with QwenAD-RFT. The whole training procedures are implemented using the ms-swift~\cite{zhao2024swiftascalablelightweightinfrastructure} framework.

\begin{figure}[t]
  \centering
  \vspace{-.5em}
  \begin{minipage}[t]{0.48\linewidth}
    \centering
    \includegraphics[width=\linewidth]{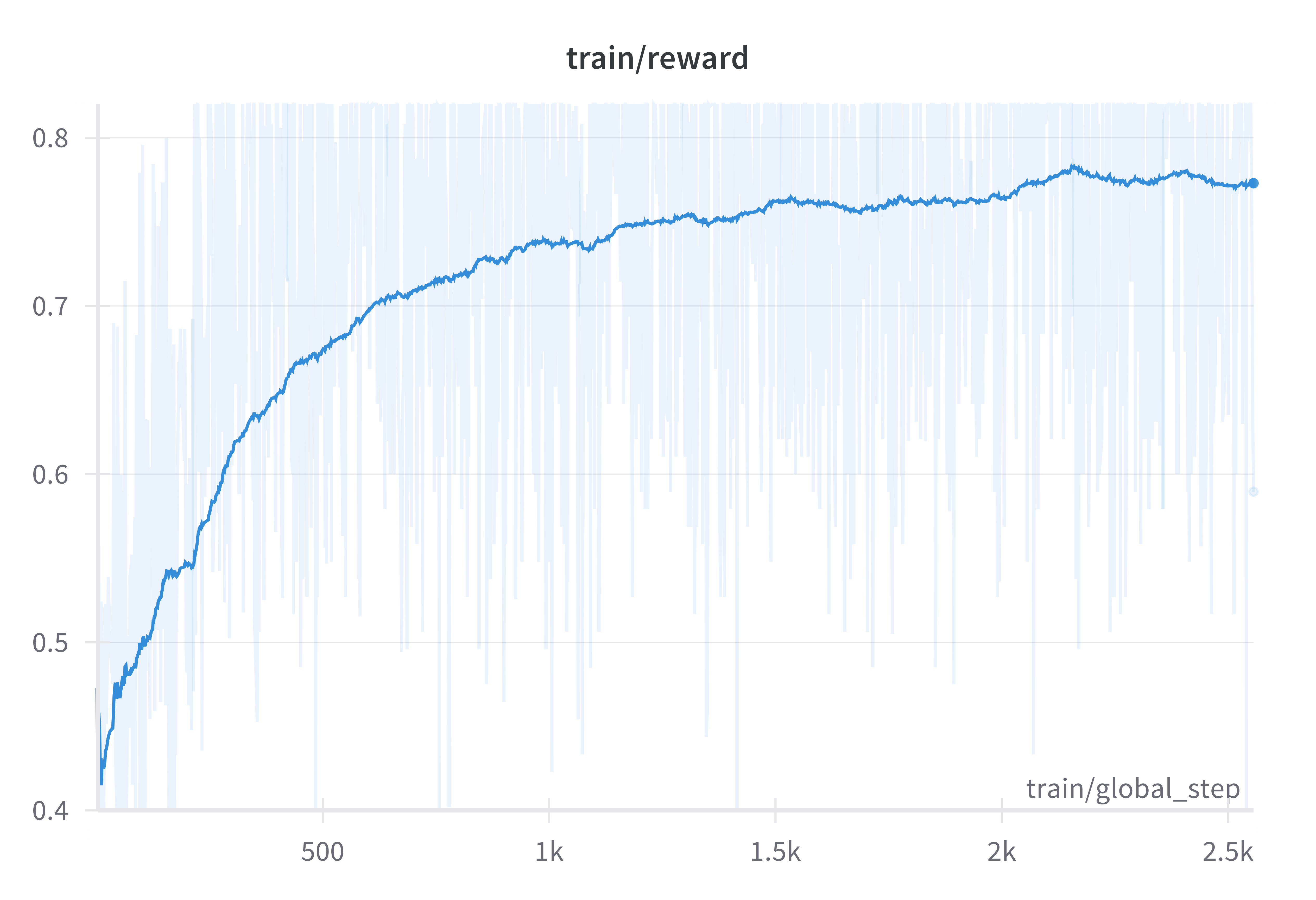}
    \caption{The variation of ove-rall rewards during training.}
    \label{fig:img1}
  \end{minipage}
  \hfill
  \begin{minipage}[t]{0.48\linewidth}
    \centering
    \includegraphics[width=\linewidth]{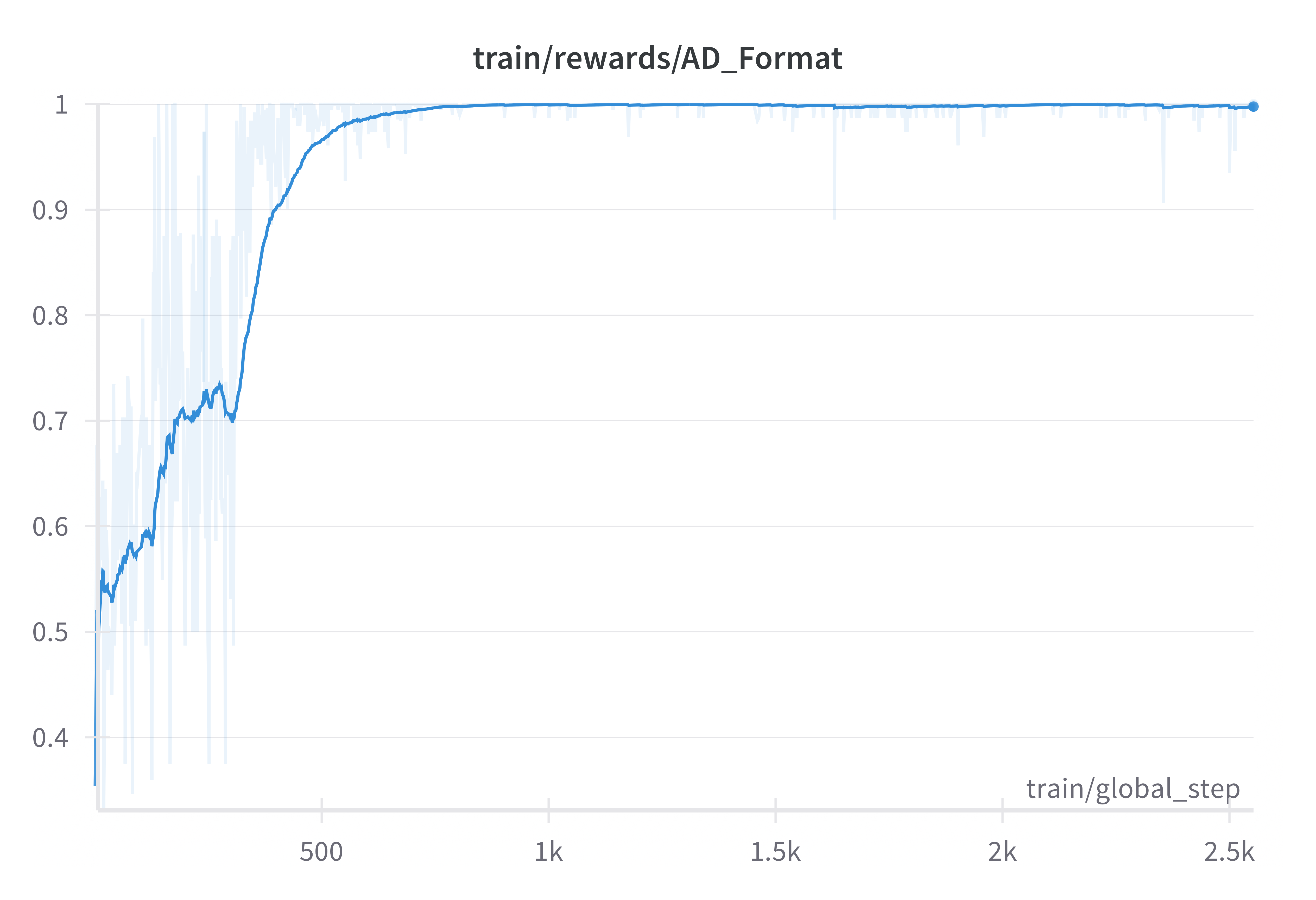}
    \caption{The variation of format rewards during training.}
    \label{fig:img2}
  \end{minipage}
\end{figure}

\subsection{Performance of QwenAD-Series}
\label{subsec:AD_eval}

\textbf{Results on MARS-Bench.}
From Table~\ref{tab:bench}, we have the following observations. \textbf{(1)} Our QwenAD series models, particularly QwenAD-SFT-RFT and QwenAD-RFT, demonstrate substantial improvements in assumptive reasoning capabilities. Notably, QwenAD-SFT-RFT, with 7B parameters, achieves an assumptive reasoning accuracy of 495.2, outperforming the baseline model (Qwen2.5-VL-7B) by 76.6 points (from 418.6). This indicates the overall effectiveness of our proposed AD approach. \textbf{(2)} When comparing three variants of QwenAD, QwenAD-SFT-RFT generally outperforms QwenAD-RFT, which suggests that SFT helps constrain the model's search space, thereby enhancing the search efficiency of RFT. 

\noindent\textbf{Results on General Benchmarks.}
We also evaluate our QwenAD series models across a wide range of multimodal comprehension benchmarks. "As presented in Table~\ref{tab:vqa}, both QwenAD-SFT and QwenAD-SFT-RFT underperform relative to the baseline model (i.e., Qwen2.5-VL~\cite{Qwen2.5-VL}), whereas QwenAD-RFT outperforms it. Considering this observation alongside the fact that a large portion of the training data is oriented toward assumptive reasoning tasks, we infer that the token-level mimicking behavior of SFT overly constrains the search space, biasing it toward the distribution of pre-defined training data and thereby limiting generalization to broader scenarios. In contrast, directly applying RFT to the base model (i.e., QwenAD-RFT) enables more effective exploration within a more optimal search space.

\subsection{Comprehensive Analysis}
\label{subsec:compr_ana}

\noindent \textbf{Effectiveness of the proposed AD-RFT reward.} 
Firstly, we visualize the variation of overall and format rewards in Fig.\ref{fig:img1} and Fig.\ref{fig:img2}, respectively. As the model is trained for only one epoch, all samples are encountered for the first time, eliminating the risk of overfitting. The steadily rising rewards illustrated in Fig.\ref{fig:img1} and Fig.\ref{fig:img2} demonstrate that our proposed AD-RFT framework is well-suited for training on data that includes both reflective (thinking) and intuitive responses.

Furthermore, we compare the performance of our proposed AD-RFT reward with the vanilla GRPO reward. As shown in Table~\ref{tab:abl}, the AD-RFT reward consistently outperforms the vanilla reward across most benchmarks, indicating that our difficulty-driven, divide-and-conquer reward design is more effective in enhancing model robustness across both simple and complex tasks.

\begin{table}[t]
\centering
\scalebox{0.95}{
\begin{tabular}{l|cccc}
\toprule
\multicolumn{1}{c}{\multirow{1}{*}{Method}} & Thk.\%Basic $\downarrow$ & Ans.\%Assum. $\downarrow$\\ 
\midrule
Qwen2.5-VL-7B     & 7.1\% & 67.3\%  \\
\midrule
\ + vanilla GRPO  & 93.1\% & 0.3\% \\
\midrule
\ + AD-SFT        & 2.3\% & 4.4\% \\
\ + AD-(SFT+GRPO) & 1.7\% & 4.1\% \\
\ + AD-GRPO       & 2.1\% & 5.3\% \\
\bottomrule
\end{tabular}
}
\caption{\textbf{Impact of reasoning token on problem difficulty determination.} Here, Thk.\%Basic, Ans.\%Assum. indicate the ratio of performing thinking when faced with basic questions, as well as directly generate answers when confronted with assumptive questions, respectively.}
\label{tab:ablation2}
\vspace{-1em}
\end{table}

\noindent \textbf{Probing of Active Deduction Behaviors.} To further inspect the model's active deduction behaviors when faced with different difficulties of questions, we calculate the behavior error rates for both basic and assumptive questions on the MARS-Bench as shown in Tab.\ref{tab:ablation2}. It can be observed that compared with vanilla GRPO reward, which compels the model to adopt long CoT reasoning (as evidenced by 93.1\% of responses exhibiting reflective behavior on basic questions), our proposed AD methods enable the model to adaptively select behavior patterns according to the varying difficulty levels of questions. 

\section{Conclusion}
\label{sec:conclu}

In conclusion, we presented MARS-Bench, a benchmark targeting Assumptive Reasoning in Multimodal Large Language Models (MLLMs), and introduced the Active Deduction (AD) method to enhance these models' reasoning capabilities. Our findings show that current MLLMs struggle with systematic reasoning problems like assumptive reasoning. Besides, Active Deduction substantially improves MLLMs' performance on assumptive tasks by guiding structured, stepwise deductive reasoning without sacrificing performance on simpler queries. This work underscores the limitations of empirical reasoning in current MLLMs and suggests a potential approach for fostering more human-like reasoning in systematic, presuppositions complex scenarios.

\bibliographystyle{ACM-Reference-Format}
\bibliography{sample-base}

\end{document}